\documentclass[10pt,twocolumn,letterpaper]{article}

\usepackage[arxiv]{cvpr}      % To produce the REVIEW version
\definecolor{cvprblue}{rgb}{0.21,0.49,0.74}
\usepackage[pagebackref,breaklinks,colorlinks,allcolors=cvprblue]{hyperref}
\usepackage{multirow}
\usepackage[table]{xcolor}
\usepackage{array}

\usepackage[section]{placeins} % 导言区
\usepackage{makecell}
\usepackage{tabularx}
\usepackage{url} 
\usepackage{bm}
\usepackage{amssymb}
\usepackage{pifont}
\usepackage{graphicx} % 导言区加一次即可
\usepackage{csquotes}
\newcommand{\cmark}{\ding{51}} % ✓
\newcommand{\xmark}{\ding{55}} % ✗
\def\paperID{N/A} % *** Enter the Paper ID here
\def\confName{CVPR}
\def\confYear{2026}
\NewDocumentCommand{\figref}{m o}{%
	Fig.~\hyperref[#1]{\ref*{#1}\IfValueT{#2}{(#2)}}%
}
\title{FUSAR-GPT: A Spatiotemporal Feature-Embedded and Two-Stage Decoupled Visual Language Model for SAR Imagery}

\author{First Author\\
Institution1\\
Institution1 address\\
{\tt\small firstauthor@i1.org}
\and
Second Author\\
Institution2\\
First line of institution2 address\\
{\tt\small secondauthor@i2.org}
}

\author{
Xiaokun Zhang$^{\dag}$,  Yi Yang$^{\dag}$$^{*}$, Ziqi Ye$^{\dag}$, Baiyun, Xiaorong Guo, \\Qingchen Fang, Ruyi Zhang, Xinpeng Zhou, Haipeng Wang$^{\ddag}$\\
\vspace{0.5em}
  {\small\linespread{0.9}\selectfont  % 左花括号开启格式作用域
    $^{\dag}$ These authors contributed equally to this work. \\
    $^{\ddag}$ Corresponding author: Haipeng Wang(hpwang@fudan.edu.cn) \\
    $^{*}$ Project lead: Yi Yang(yangy22@m.fudan.edu.cn)
  }  % 
  }

\author{
  \textbf{Xiaokun Zhang}$^{*}$, \quad
  \textbf{Yi Yang}$^{*\dag}$, \quad
  \textbf{Ziqi Ye}$^{*}$, \quad
  \textbf{Baiyun}, \quad
  \textbf{Xiaorong Guo}, \\
  \textbf{Qingchen Fang}, \quad
  \textbf{Ruyi Zhang}, \quad
  \textbf{Xinpeng Zhou}, \quad
  \textbf{Haipeng Wang}$^{\ddag}$
  \\[1.8ex]
  \small $^{*}$Equal contribution \quad 
  \small $^{*\dag}$Project lead: Yi Yang (\href{mailto:yangy22@m.fudan.edu.cn}{\color{blue}{yangy22@m.fudan.edu.cn}}) \\
  \small $^{\ddag}$Corresponding author: Haipeng Wang (\href{mailto:hpwang@fudan.edu.cn}{\color{blue}{hpwang@fudan.edu.cn}}) \\
  \small Fudan University
}

\author{
  \textbf{Xiaokun Zhang}$^{1*}$, \quad
  \textbf{Yi Yang}$^{1*\dag}$, \quad
  \textbf{Ziqi Ye}$^{1,2*}$, \quad
  \textbf{Baiyun}$^{1}$, \quad
  \textbf{Xiaorong Guo}$^{1}$, \\
  \textbf{Qingchen Fang}$^{1}$, \quad
  \textbf{Ruyi Zhang}$^{1}$, \quad
  \textbf{Xinpeng Zhou}$^{1}$, \quad
  \textbf{Haipeng Wang}$^{1\ddag}$
  \\[1.8ex]
  \small $^{1}$Discipline and Technology Center of Microwave Vision Intelligent Sensing, Fudan University \quad $^{2}$Shanghai Innovation Institute \\
  \small $^{*}$Equal contribution \quad 
  \small $^{\dag}$Project leader: Yi Yang (\href{mailto:yangy22@m.fudan.edu.cn}{\color{blue}{yangy22@m.fudan.edu.cn}}) \\
  \small $^{\ddag}$Corresponding author: Haipeng Wang (\href{mailto:hpwang@fudan.edu.cn}{\color{blue}{hpwang@fudan.edu.cn}})
}

\begin{document}
\maketitle
\begin{abstract}
\hspace{2em}Research on the intelligent interpretation of all-weather, all-time Synthetic Aperture Radar (SAR) is crucial for advancing remote sensing applications. In recent years, although Visual Language Models (VLMs) have demonstrated strong open-world understanding capabilities on RGB images, their performance is severely limited when directly applied to the SAR field due to the complexity of the imaging mechanism, sensitivity to scattering features, and the scarcity of high-quality text corpora. To systematically address this issue, we constructed the inaugural SAR Image-Text-AlphaEarth feature triplet dataset and developed FUSAR-GPT, a VLM specifically for SAR. FUSAR-GPT innovatively introduces a geospatial baseline model as a \enquote{world knowledge}  prior and embeds multi-source remote-sensing temporal features into the model's visual backbone via \enquote{spatiotemporal anchors}, enabling dynamic compensation for the sparse representation of targets in SAR images. Furthermore, we designed a two-stage SFT strategy to decouple the knowledge injection and task execution of large models. The spatiotemporal feature embedding and the two-stage decoupling paradigm enable FUSAR-GPT to achieve state-of-the-art performance across several typical remote sensing visual-language benchmark tests, significantly outperforming mainstream baseline models by over 10\%.
\end{abstract}    
\section{Introduction}
\label{sec:intro}
 The intelligent interpretation of SAR imagery has long been recognized as a pivotal research focus in the field of remote sensing\cite{11173694}. 
 Despite progress in specific tasks like object detection, SAR image interpretation remains highly challenging due to its complex electromagnetic scattering and coherence effects. Concurrently, multimodal language models (e.g., CLIP \cite{pmlr-v139-radford21a}, BLIP \cite{pmlr-v162-li22n}) have excelled in natural image domains, demonstrating strong semantic alignment and cognitive reasoning by pre-training on large-scale image-text pairs.

\begin{figure}[tbp]
\centerline{\includegraphics[width=0.5\textwidth]{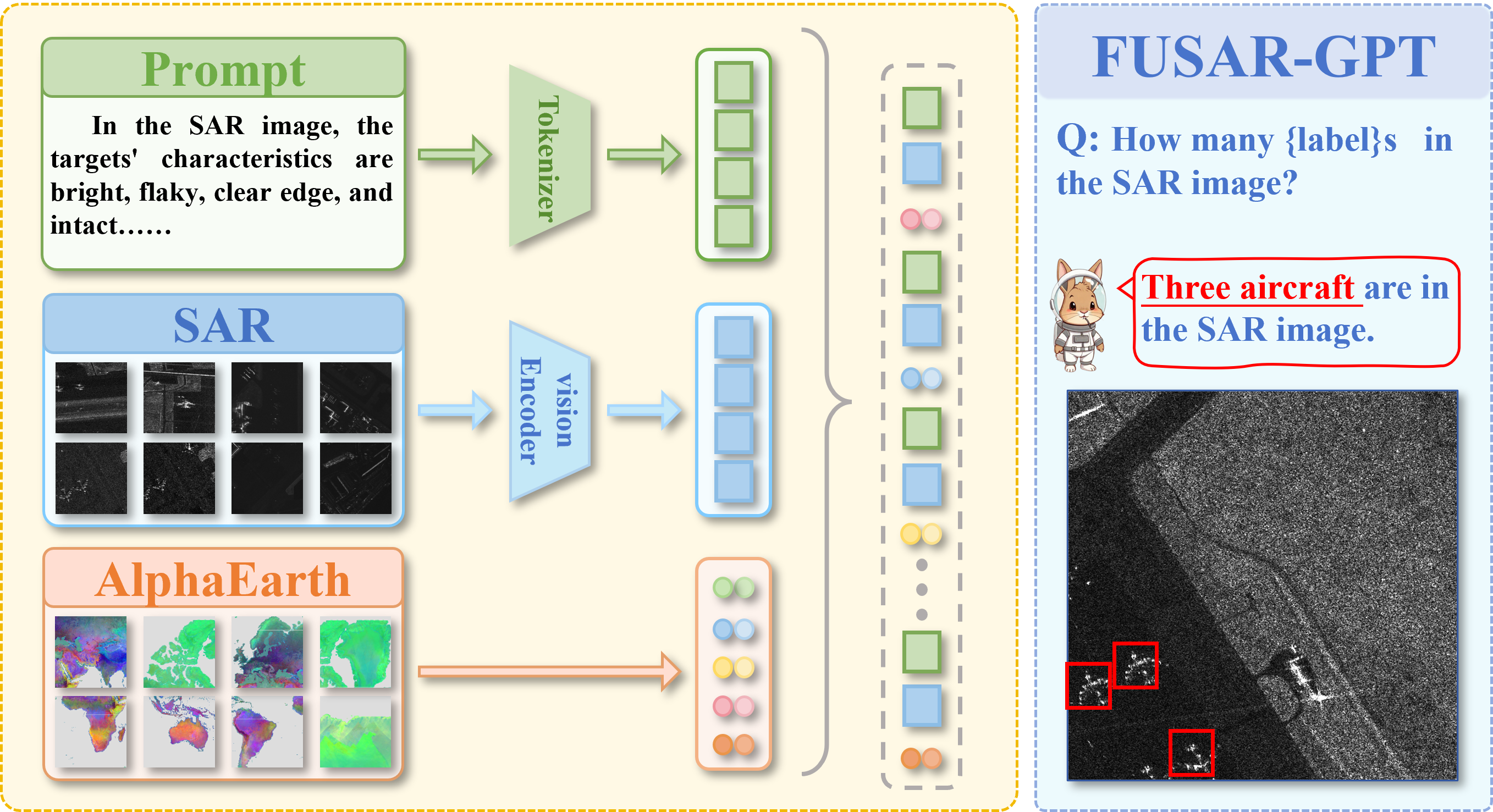}}
\caption{
FUSAR-GPT: Embeds spatiotemporal features for SAR feature compensation and uses two-stage decoupled SFT for target interpretation.}
\label{fig1}
\end{figure}

This technological trend is gradually extending into the field of remote sensing, with several studies attempting to adapt vision-language models to remote sensing image scenarios, demonstrating their feasibility on optical remote sensing images. However, existing approaches predominantly rely on RGB imagery and have not yet accounted for the modality-specific characteristics and semantic complexity inherent in SAR images \cite{kuckreja2024geochat}. Achieving this objective requires overcoming several key challenges:

% \begin{figure}[tbp]
% \centerline{\includegraphics[width=0.4\textwidth]{fig2.png}}
% \caption{Challenges in developing SAR visual language models.}
% \label{fig2}
% \end{figure}

\noindent

\textbf{(a) SAR-Optical Modal Difference: }Existing multimodal models, pretrained on large-scale visible light datasets, possess feature representations that fundamentally mismatch the data distribution of SAR imagery due to its distinct imaging mechanism. Directly transferring these models fails to adapt to this inter-modal discrepancy\cite{gao2025combining}, resulting in poor generalization and limited interpretive performance. This ultimately renders the transfer paradigm ineffective for SAR tasks.

\noindent

\textbf{(b) Neglecting Geospatial Priors: }Geographic information serves as an irreplaceable strong prior constraint for remote sensing interpretation. Current SAR interpretation research, however, suffers from a systematic flaw: it adopts technical frameworks designed for natural images, resulting in models that lack spatial awareness. As a result, essential geographic scene priors are underutilized, weakening high-level semantic reasoning in vision–language models (e.g., confusing urban buildings with metallic objects) and exacerbating hallucinations.
% This leads to the neglect of crucial geographic scene priors—a failure that causes vision-language models to lose high-level cognitive reasoning capabilities (e.g., inability to distinguish city buildings from metal tools) and contributes to model hallucination.

\noindent

\textbf{(c) Information Sparsity: } Owing to the coherent imaging mechanism of SAR and its extreme sensitivity to geometric and dielectric properties, SAR images inherently exhibit an extremely high dynamic range and information sparsity. This polarized data distribution—where artificial targets (such as corner reflectors) generate excessively saturated strong scattering, and natural objects (such as water surfaces) appear as extensive dark regions—causes the model’s attention to be dominated by a few bright pixels. Consequently, the rich contextual semantics embedded in the vast dark areas are systematically overlooked, severely limiting the VLM's ability to achieve comprehensive semantic comprehension of the scene.

To address the aforementioned issues, we propose FUSAR-GPT. Built upon Qwen2.5-VL-7B, FUSAR-GPT introduces a spatiotemporal semantic embedding module that incorporates geospatial features as \enquote{world knowledge} priors into the model’s visual encoder, dynamically compensating for the sparse representation of targets in SAR images. Furthermore, we develop a two-stage supervised fine-tuning (SFT) strategy to decouple knowledge injection from task execution. The main contributions of this paper are summarized as follows:

\begin{itemize}
    \item This study establishes the first \enquote{SAR Image-Text-Feature} triplet data paradigm, innovatively introducing geospatial  features as a third modality. By leveraging spatiotemporal anchors, this paradigm integrates rich world knowledge priors to enable dynamic semantic compensation for sparse SAR features.
    \item We propose the Token-wise Linear Modulation (TLM) fusion module. TLM achieves fine-grained, dynamic semantic injection by employing local spatial alignment and per-channel linear modulation to transform high-dimensional priors into spatially differentiated parameters for visual token adjustment.
    \item We designed an innovative two-stage SFT paradigm to systematically decouple SAR modality knowledge injection (establishing cognitive capability) from the execution of downstream tasks (endowing higher-level analytical and reasoning abilities).
    \item FUSAR-GPT achieved state-of-the-art performance in multiple SAR interpretation tasks. It demonstrated a significant performance improvement of more than 10\% over mainstream VLMs.
\end{itemize}
%-------------------------------------------------------------------------

\section{Related Work}
\label{sec:relatework}

%-------------------------------------------------------------------------
\subsection{Research on SAR image interpretation}
Early SAR image processing relied on statistical modeling and manual feature extraction for tasks like speckle suppression and basic recognition \cite{9568916}. More recently, the explosive development of deep learning, particularly CNN-based frameworks, has delivered significant performance improvements in specific tasks like SAR target detection\cite{liu2025atrnet}, semantic segmentation\cite{sun2026fully}, and scene classification \cite{11144037}, \cite{9554697}, \cite{9864735}, \cite{8633175}.

Specialized techniques have also emerged: SARDet-CL, a contrastive learning framework, improves detection with limited labeled samples by adaptively enhancing features through dynamic quantization and imaging-mechanism constraints \cite{yang2025sardet}. Furthermore, to address geometric challenges like large-scale scenes, progressive detection frameworks integrate traditional methods with deep learning to achieve high efficiency and accuracy \cite{jhc}.

Despite these advancements, deep learning methods remain limited to modeling low-level visual features, leading to a serious deficiency in high-dimensional semantic understanding and cognitive-level reasoning. Therefore, a new interpretation paradigm is urgently required to bridge the SAR vision and language modalities and enable semantic knowledge injection.

%-------------------------------------------------------------------------
\subsection{Research on Vision-Language Models in Remote Sensing}
General VLMs have made substantial breakthroughs in the natural image domain\cite{huang2024unpaired}\cite{zhou2024towards}\cite{li2025star}\cite{liu2012texture}, rapidly driving the paradigm shift of remote sensing interpretation toward a cognitive level. Researchers in remote sensing aim to bridge the semantic gap between remote sensing images and natural language by constructing large-scale image–text alignment datasets and developing generative models for data augmentation \cite{ye2025object} \cite{Liu2025Causal}. Representative works include EarthGPT \cite{10547418}, which proposed a general multi-sensor image understanding framework emphasizing visual perception and cross-modal mutual understanding; GeoChat \cite{kuckreja2024geochat} developed the first multi-task conversational remote sensing VLM, focusing on region-level reasoning for high-resolution images. EarthDial \cite{11093668} further expanded its scope to support interactive dialogues across multi-temporal and multi-sensor data (including SAR), and released a large-scale instruction dataset. In addition, works such as SkyScript \cite{Wang_Prabha_Huang_Wu_Rajagopal_2024} and RemoteClip\cite{liu2024remoteclip} have also made significant progress in tasks including VQA, description generation, and functional area recognition for optical remote sensing images through large-scale datasets and instruction fine-tuning.

However, despite the rapid development of remote sensing VLM\cite{geor1}\cite{11093668}\cite{earthgptx}, existing research has mainly focused on visible (RGB) or multispectral images, and studies specifically targeting visual-language models for SAR images remain at an early stage and are extremely scarce.

Only a few preliminary works (e.g., SARCLIP \cite{JIANG202617}, SARLANG-1M \cite{wei2026sarlang}) have been attempted, primarily achieving basic modal alignment and zero-shot generalization. However, their capabilities remain confined to feature-level alignment, critically failing to account for SAR's intrinsic modality specificity and systematically overlooking crucial geospatial prior information. Consequently, advancing intelligent SAR interpretation toward multimodal, open-ended understanding remains highly challenging.

\section{Method}

\begin{figure*}[tbp]
\centering
\includegraphics[width=\textwidth]{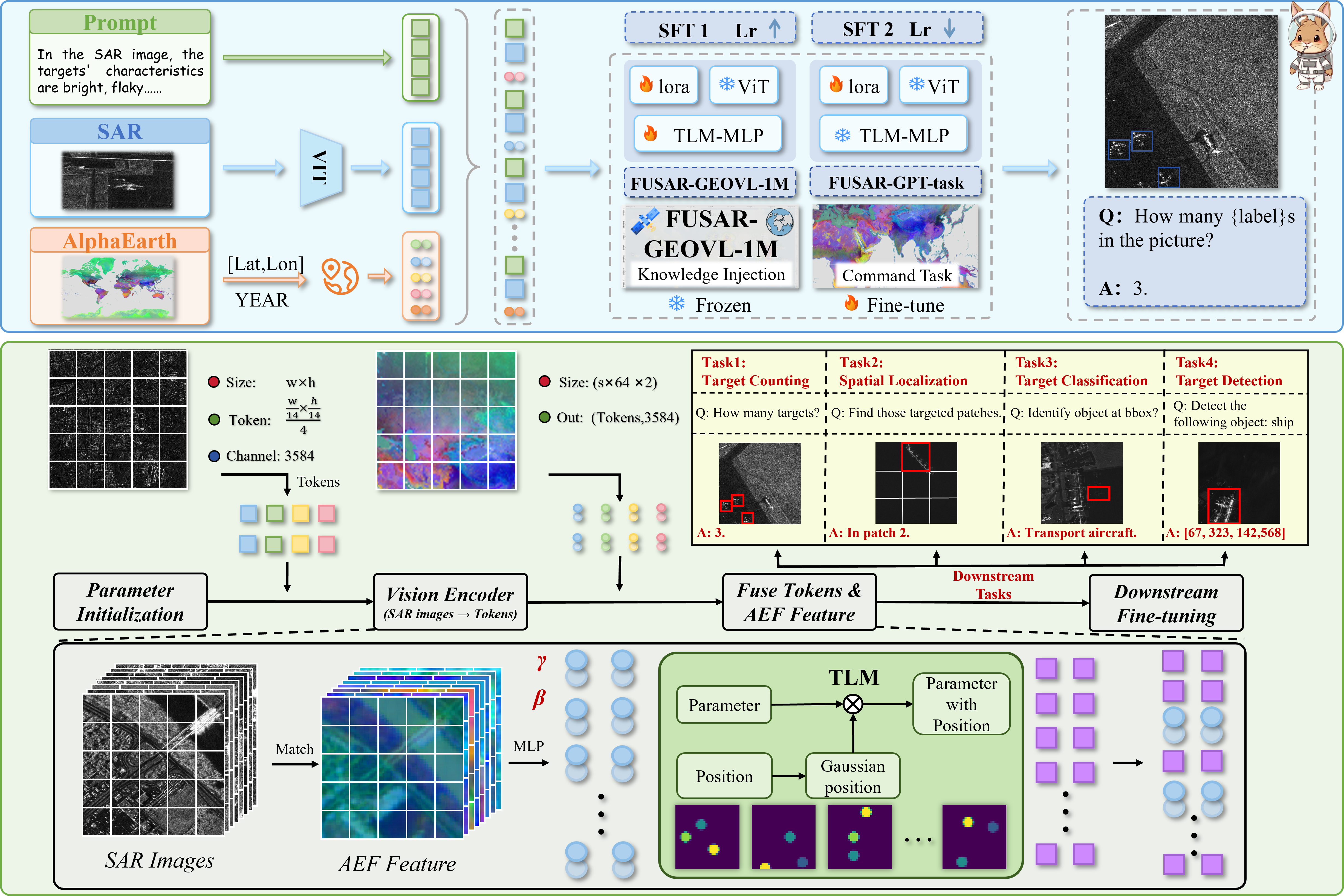}
\caption{Overview of FUSAR-GPT. The framework adopts a two-stage training strategy: Stage-1 jointly updates LoRA and the TLM-MLP for multimodal prior injection, while Stage-2 fine-tunes only LoRA for task adaptation. The TLM module fuses SAR visual tokens with AEF priors by generating spatially informed modulation parameters, which are applied to the visual tokens to enhance downstream reasoning.}
\label{fig3}
\end{figure*}

FUSAR-GPT is a SAR vision-language model, built as a customized enhancement upon the Qwen2.5VL-7B \cite{qwen25vl} architecture (overall framework shown in \figref{fig3}). Its core design encompasses two aspects: multi-source remote sensing temporal feature embedding, and a two-stage decoupled SFT strategy.

\subsection{AlphaEarth Multi-Source Temporal Feature Extraction}

The microwave active imaging mechanism of SAR leads to severe information imbalance and semantic sparsity in image content. To address this bottleneck, FUSAR-GPT introduces AEF \cite{alphaearthfoundations} as a multi-source temporal feature compensation. AEF is a global remote sensing foundation model that integrates heterogeneous multi-source data, such as optical, SAR, and LiDAR, into a 64-dimensional continuous spatiotemporal embedding field. This embedding field provides FUSAR-GPT with the cross-modal prior knowledge necessary to compensate for sparse SAR features and achieve robust understanding.

To precisely align the multi-source prior knowledge of AEF with the visual features of SAR images, we design a feature extraction process based on spatio-temporal anchors.

First, for any given SAR image, we determine its spatio-temporal bounding box $B$.
\begin{equation}
B = [lon_{\min}, lat_{\min}, lon_{\max}, lat_{\max}, y]
\label{eq:bbox}
\end{equation}

Where $[lon,lat]$ specifies the geographic coverage of the image and $y$ denotes its imaging year.

Subsequently, to capture the AEF continuous embedding field, we construct an $N_{lon} \times N_{lat}$ regular sampling grid over $B$. The geographic coordinates $(lon_i, lat_j)$ for node $(i, j)$ are computed as:
\begin{equation}
\begin{gathered}
    \Delta lon = \frac{lon_{\max} - lon_{\min}}{N_{lon} - 1}; \quad \Delta lat = \frac{lat_{\max} - lat_{\min}}{N_{lat} - 1} \\
    lon_i = lon_{\min} + i \cdot \Delta lon; \quad lat_j = lat_{\min} + j \cdot \Delta lat
\end{gathered}
\label{eq:grid_all}
\end{equation}
where $0 \le i < N_{lon}$ and $0 \le j < N_{lat}$.

On each geographic node $(lon_i,lat_j)$, we query the AEF model $I_i$ to obtain the 64-dimensional embedding vector corresponding to year $y$:$a_{ij} = I_y(lon_i, lat_j) \in \mathbb{R}^{64}$.

Crucially, to align AEF semantic features with SAR visual features, we perform a linear mapping from each geographic coordinate $(lon_i,lat_j)$ to the corresponding pixel coordinate system $(x_{ij},y_{ij})$ of the SAR image.

Ultimately, the extracted AEF features are organized into a set $\mathcal{F}$, which encapsulates the precise alignment relationships among geolocation, pixel indexing, and multi-source semantic embeddings:

\begin{equation}
\mathcal{F}(B, y) = \{ (lon_{ij}, lat_{ij}, x_{ij}, y_{ij}, a_{ij}) \}_{i,j}
\label{eq:feature_set}
\end{equation}

This process ensures that cross-modal prior knowledge (AEF) can be accurately injected into specific spatial locations within SAR images.

\begin{figure*}[tbp]
\centerline{\includegraphics[width=\textwidth]{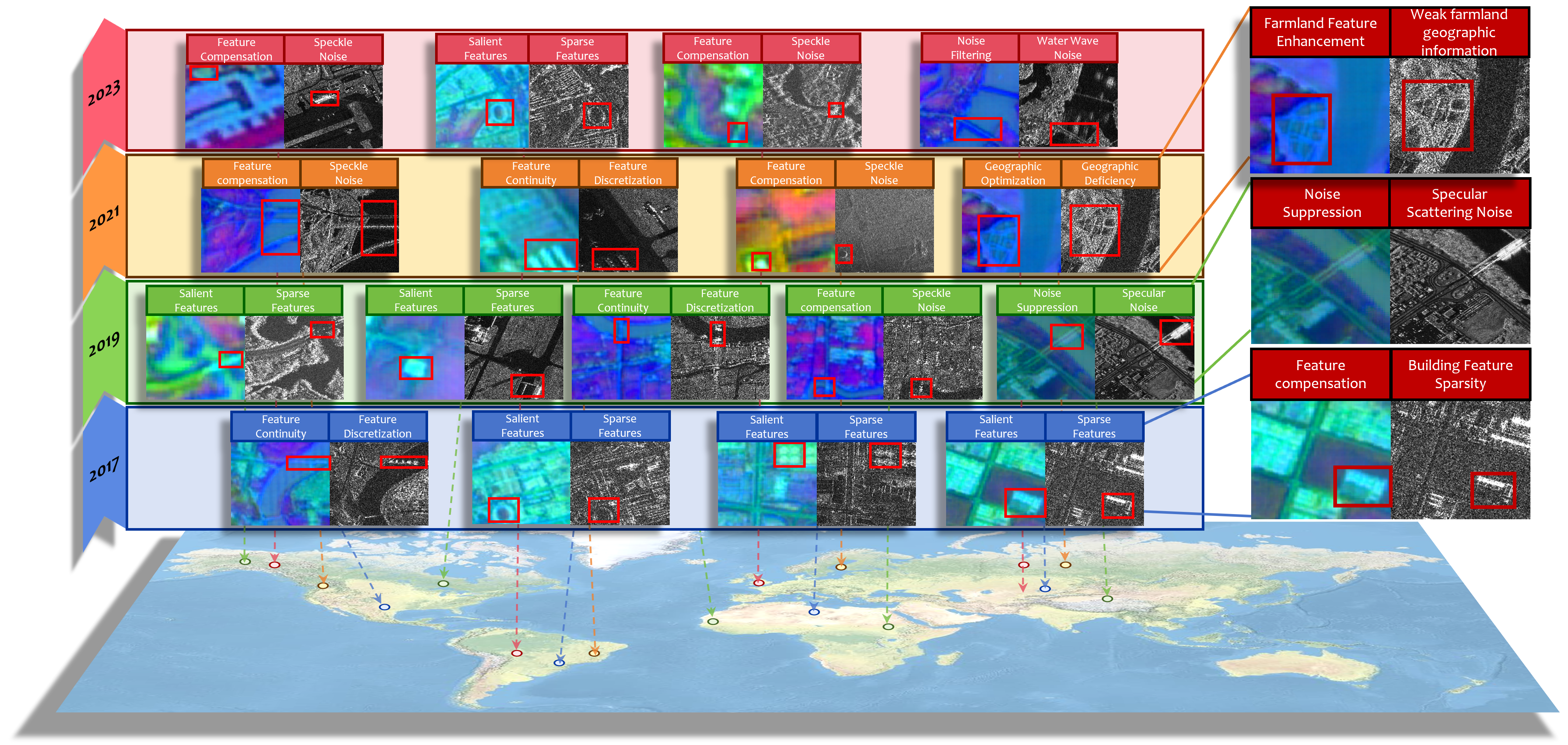}}
\caption{AEF–SAR Visual Comparison. AEF embeddings extracted at different times are visualized by mapping channels 1, 16 and 9 of the 64-dimensional feature vector to RGB. }
\label{fig4}
\end{figure*}

\figref{fig4} verifies the significant complementarity between SAR and AEF features. Leveraging spatiotemporal anchor alignment, AEF provides critical dynamic semantic compensation for the sparse and incomplete SAR representations caused by imaging limitations. As shown in the enlarged area on the right side of \figref{fig4}, AEF enhances weak features (such as farmland areas) and provides clean, stable object representations even in areas affected by specular scattering noise.

Although AEF provides multi-source temporal embeddings, we do not explicitly model temporal dynamics. 
Instead, AEF is used as a stable geospatial prior, as macroscopic geographic semantics remain relatively consistent at the annual scale, providing consistent large-scale context.
% To reduce potential inconsistencies arising from temporal mismatch, the TLM module performs spatially adaptive fusion via Gaussian-weighted modulation.

\subsection{TLM fusion module}

% The analysis in the previous section intuitively verifies the significant complementarity of AEF features, confirming their ability to provide critical dynamic semantic compensation for SAR images. However, a core challenge arises: how can this external multi-source prior knowledge 

% These two types of representations are heterogeneous in both modality and structure: AEF is represented as discrete sampling points over a continuous geographic field, whereas the visual tokens are deep features extracted from image patches by the backbone network. If a simple feature-concatenation or summation strategy is used, it would not only incur substantial computational cost and pose alignment difficulties, but also easily interfere with the SAR spatial representations already learned by the backbone visual pathway.

% To address this, FUSAR-GPT introduces the Token-wise Linear Modulation (TLM) fusion module. The design is inspired by conditional normalization techniques.TLM does not treat AEF as “data” to be processed, but rather as a “condition” that is used to dynamically generate a set of modulation parameters $(\bm{\gamma}, \bm{\beta})$.This set of parameters then applies a channel-wise affine transformation to the SAR visual tokens $X$.In this way, the semantic priors of AEF can guide, rather than interfere with, the fine adjustment of SAR visual representations across both spatial and channel dimensions.
The previous analysis demonstrates that AEF provides complementary semantic cues for SAR imagery, supplying dynamic and globally consistent priors. The remaining challenge is how to efficiently introduce this external multi-source knowledge into the visual representation space.

AEF and visual tokens are heterogeneous in both form and modality: AEF consists of sparsely sampled geo-semantic vectors, while visual tokens are dense deep features extracted from image patches. Naive strategies such as concatenation or direct fusion cause alignment mismatches, increase computation, and may distort the spatial structure already learned by the visual backbone.

To address this, FUSAR-GPT adopts the TLM module, inspired by conditional normalization. Instead of treating AEF as additional feature inputs, TLM interprets them as conditioning signals that generate modulation parameters $(\bm{\gamma}, \bm{\beta})$, which apply channel-wise affine transformations to the visual tokens $X$. Through this lightweight modulation, AEF guides the refinement of SAR representations without disrupting the backbone’s spatial encoding.

\textbf{TLM for AEF–SAR Fusion.}
After obtaining the AEF feature field $\mathcal{F}(B,y)$ that is strictly aligned with the SAR image space, FUSAR\text{-}GPT introduces a TLM fusion module at the output of the visual encoder in order to explicitly inject the cross-modal prior information without disrupting the structure of the backbone visual pathway. Note that the token sequence output by the visual encoder is as follows:
$\mathbf{X} = [\mathbf{x}_1,\ldots,\mathbf{x}_T]^\top \in \mathbb{R}^{T\times C},$
where $\mathbf{x}_t\in\mathbb{R}^{C}$ denotes the channel features of the $t$-th visual token, and $C$ is the visual hidden dimension. Correspondingly, the structured priors obtained from AEF sampling are organized into a vector sequence as follows:
$\mathbf{V} = [\mathbf{v}_1,\ldots,\mathbf{v}_S]^\top \in \mathbb{R}^{S\times D},$
$\mathbf{P} = [(y_1,x_1),\ldots,(y_S,x_S)]^\top \in [0,1]^{S\times 2}, $
where $\mathbf{v}_i\in\mathbb{R}^{D}$ is the $i$-th AEF embedding vector, $(y_i,x_i)$ is its normalized spatial coordinate on the SAR image, and $S$ is the number of valid prior vectors for the sample.

\textbf{Prior-to-Modulation Mapping.}
The core idea of TLM is to treat each AEF vector $\mathbf{v}_i$ as a conditioning signal that generates channel-wise modulation parameters for the visual tokens. Specifically, $\mathbf{v}_i \in \mathbb{R}^D$ is projected into a $2C$-dimensional space to produce a pair of scaling and shifting coefficients $(\bm{\gamma}_i, \bm{\beta}_i)\in\mathbb{R}^C\times\mathbb{R}^C$ for modulating the visual feature $\mathbf{x}_t$.
Specifically, a two-layer MLP with SiLU activation is employed:
\begin{equation}
\mathbf{h}_i = \phi(\mathbf{W}_1 \mathbf{v}_i + \mathbf{b}_1) , (\bm{\gamma}_i,\bm{\beta}_i) = \mathbf{W}_2 \mathbf{h}_i + \mathbf{b}_2
\end{equation}
% \begin{equation}

% \label{eq:film-mlp}
% \end{equation}

Where $\phi(\cdot)$ is the SiLU activation function, $\mathbf{W}_1\in\mathbb{R}^{h\times D}$, $\mathbf{W}_2\in\mathbb{R}^{2C\times h}$, and $h$ is the hidden dimension.
Thus, the modulation coefficient matrices corresponding to all external vectors can be obtained as
% $
% \Gamma=\begin{bmatrix}\bm{\gamma}_1 & \bm{\gamma}_2 & \cdots & \bm{\gamma}_S\end{bmatrix}\in\mathbb{R}^{C\times S},
% B=\begin{bmatrix}\bm{\beta}_1 & \bm{\beta}_2 & \cdots & \bm{\beta}_S\end{bmatrix}\in\mathbb{R}^{C\times S}.
% $
\begin{equation}
\Gamma = 
\begin{bmatrix}
\bm{\gamma}_1^\top \\
\vdots \\
\bm{\gamma}_S^\top
\end{bmatrix}
\in\mathbb{R}^{S\times C},
B = 
\begin{bmatrix}
\bm{\beta}_1^\top \\
\vdots \\
\bm{\beta}_S^\top
\end{bmatrix}
\in\mathbb{R}^{S\times C} 
\end{equation}

\textbf{Local Spatial Alignment Based on Gaussian Weights.}
% The above steps generate the $S$ groups of modulation parameters $(\Gamma, B)$.However, these parameters are defined on the $S$ sparse AEF geographic locations $\mathbf{P}$, while our goal is to modulate the dense visual tokens $\mathbf{X}$. Therefore, we must establish a spatial alignment between the two and interpolate the modulation parameters from the sparse AEF positions onto the dense visual grid $(h,w)$.To achieve fine alignment between the prior information and the visual tokens, the TLM module utilizes the spatial mesh information produced in the intermediate layers of the visual encoder.To achieve fine alignment between the prior information and the visual tokens, the TLM module utilizes the spatial mesh information produced in the intermediate layers of the visual encoder.The normalized coordinates in $P$ are mapped onto the actual grid of the visual feature map. Let the spatial size of the feature map be $H\times W$.Then, the continuous grid coordinates of the $i$ prior vector are as follows:
The above procedure produces $S$ sets of modulation parameters $(\Gamma, B)$ defined on the sparse AEF locations $\mathbf{P}$. To modulate the dense visual tokens $\mathbf{X}$, these parameters must be spatially aligned and interpolated onto the visual grid. TLM achieves this by leveraging the intermediate spatial mesh of the visual encoder. The normalized AEF coordinates are projected onto the feature map of size $H\times W$, yielding the continuous grid location of the $i$-th prior vector:
\begin{equation}
\tilde{y}_i = y_i (H-1) , \tilde{x}_i = x_i (W-1)
\end{equation}

At each spatial position $(h,w)$, we use a distance-based Gaussian kernel to define the interpolation weights, and define an unnormalized Gaussian weight for the $i$-th external vector as follows:
\begin{equation}
\tilde{w}_i(h,w) =
\exp\!\Big(
-\frac{(h-\tilde{y}_i)^2 + (w-\tilde{x}_i)^2}{2\sigma^2}
\Big)
\label{eq:gaussian-weight}
\end{equation}

Subsequently, column-wise normalization is performed along the prior-vector dimension so that the weights at each spatial position sum to 1:
\begin{equation}
w_i(h,w) =
\frac{\tilde{w}_i(h,w)}{\sum_{j=1}^{S} \tilde{w}_j(h,w) + \varepsilon} , \sum_{i=1}^{S} w_i(h,w) = 1
\end{equation}

By flattening the two-dimensional index $(h,w)$ into a single index $k = hW + w$,
the overall weight matrix can be obtained as follows:
\begin{equation}
W = [w_i(k)]_{i,k}\in\mathbb{R}^{S\times HW}
\end{equation}

The $k$-th column corresponds to the normalized weights contributed by each external vector to the $k$-th spatial location on the feature map.

\textbf{TLM Fusion and Token Alignment.}
After obtaining the spatial interpolation weight matrix $W$,
we apply weighted aggregation to the modulation parameters $(\Gamma, B)$ defined at the sparse AEF positions. In this way, we obtain a channel modulation field that is strictly aligned with the visual feature map:
% \begin{equation}
% \Gamma_{\mathrm{hw}} = W^\top \Gamma \in \mathbb{R}^{HW\times C}
% \end{equation}
% \begin{equation}
% B_{\mathrm{hw}} = W^\top B \in \mathbb{R}^{HW\times C}
% \label{eq:film-hw}
% \end{equation}

\begin{equation}
\Gamma_{\mathrm{hw}},\; B_{\mathrm{hw}} = W^\top (\Gamma,\; B) \in \mathbb{R}^{HW \times C}
\end{equation}

Treat $\Gamma_{\mathrm{hw}}$ and $B_{\mathrm{hw}}$ as modulation fields defined on the discrete spatial grid. They can be written respectively as $\Gamma_{\mathrm{hw}} \leftrightarrow \bm{\gamma}(h,w),$ $B_{\mathrm{hw}} \leftrightarrow \bm{\beta}(h,w),$
% \begin{equation}
% \Gamma_{\mathrm{hw}} \leftrightarrow \bm{\gamma}(h,w), 
% \end{equation}
% \begin{equation}
% B_{\mathrm{hw}} \leftrightarrow \bm{\beta}(h,w), 
% \end{equation}
where $\boldsymbol{\gamma}(h,w)$ and $\boldsymbol{\beta}(h,w)\in\mathbb{R}^{C}$ denote the channel-wise scaling and offset coefficients at the spatial location $(h,w)$. 

The final step is to apply this modulation field to the one-dimensional token sequence $\mathbf{X} \in \mathbb{R}^{T \times C}$ output by the visual encoder. 
% In the visual encoder adopted in this work, the output is the visual token sequence $\mathbf{X}\in\mathbb{R}^{T\times C}$.
% To characterize the correspondence with spatial positions, we reorganize the sequence into a set indexed by an $H \times W$ spatial grid.
We establish a bijective mapping $\pi$ from the spatial grid to the token index set, such that the $\pi(h,w)$-th element in the visual token sequence uniquely corresponds to the spatial position $(h,w)$.
% After establishing the mapping $\pi$,the modulation coefficients
% $\boldsymbol{\gamma}(h,w)$ and $\boldsymbol{\beta}(h,w)$
% defined on the grid index $(h,w)$
% can be directly associated with the corresponding positions in the visual token sequence through this mapping,
% yielding the token-aligned modulation coefficients
% $\boldsymbol{\gamma}{\pi(h,w)}$ and $\boldsymbol{\beta}{\pi(h,w)}$.

% \begin{equation}
% \bm{\gamma}_{\pi(h,w)}=\bm{\gamma}(h,w),
% \end{equation}
% \begin{equation}
% \bm{\beta}_{\pi(h,w)}=\bm{\beta}(h,w),
% \end{equation}
Finally, channel-wise TLM fusion for each visual token is achieved.
% through the mapping $t = \pi(h,w)$:
\begin{equation}
\mathbf{x}'_{\pi(h,w)}
=
\mathbf{x}_{\pi(h,w)}
\odot \bigl(1+\bm{\gamma}(h,w)\bigr)
+
\bm{\beta}(h,w)
\label{eq:film-token}
\end{equation}

This process injects AEF priors into spatial and channel dimensions without modifying the main visual pathway, effectively improving the stability and discriminability of SAR representations.
% This process enables the fine-grained injection of AEF semantic priors into both the spatial and channel dimensions without altering the main visual pathway, thereby substantially enhancing the stability and discriminability of SAR representations.

\subsection{Decoupled Two-Stage SFT}

General-purpose VLMs (such as Qwen \cite{qwen25vl} and LLaVA \cite{llava}) exhibit a substantial domain gap in SAR applications due to optical pre-training. To prevent suboptimal performance caused by the simultaneous, conflicting optimization of multimodal fusion (SAR, AEF, and text) and instruction-driven task execution in a single-stage scheme, we propose a two-stage decoupled supervised fine-tuning (SFT) framework. The datasets used in the two stages are summarized in Table \ref{tab:dataset_111}. 

\begin{table}[t]
  \centering
  \caption{Training datasets used in this work: Stage 1 fine-tuning is performed on FUSAR-GEOVL-1M, and Stage 2 fine-tuning and evaluation are conducted on FUSAR-GPT.}
  \label{tab:dataset_111}
  \huge
\resizebox{\columnwidth}{!}{
  \begin{tabular}{lllllll}
  \toprule
  \rowcolor{gray!15}
    Name & Year & Band & Satellite & Pol & AEF & Example \\
    \midrule
    \makecell{FUSAR\\-GEOVL-1M} & 2025 & C,X,Ku &
       GF3, QL1, HT1 &
      Multi & \xmark  &\makecell[l]{\raisebox{-0.3\height}{\includegraphics[width=0.3\columnwidth]{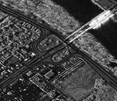}}}\\
    \rowcolor{blue!5}
    \textbf{FUSAR-GPT} & \textbf{2025} & \textbf{C,X,Ku,L} &
     \makecell[l]{\textbf{GF3, QL1} \\ \textbf{Sentinel-1A, HT1}\\ \textbf{Sentinel-1B,ALOS}} &
      \textbf{Multi} & \cmark & \makecell[l]{\raisebox{-0.3\height}{\includegraphics[width=0.3\columnwidth]{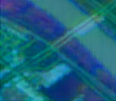}}}\\
    \bottomrule
  \end{tabular}%
          }
\end{table}

We decompose the parameter set of FUSAR-GPT, denoted as $\Theta$, into four components: $\Theta = \theta_v \cup \theta_{ae} \cup \theta_{llm}\cup \theta_{\text{lora}}$, corresponding respectively to the visual encoder $E_v$, the AEF feature-embedding MLP $E_{ae}$, the backbone large language model $L$, and the LoRA adapters injected into the language model. Our decoupling framework performs staged optimization over different subsets of parameters.

\textbf{Stage I: Cross-Modal Alignment and Knowledge Injection}
The core objective of this phase is to enable the model to understand and integrate SAR visual representations, AEF geographic priors, and textual descriptive semantics.
We perform instruction fine-tuning using the dataset $D_1 = \{(I_{sar}, F_{ae}, T_{desc})\}$, where $I_{sar}$ denotes the SAR image, $F_{ae}$ is the corresponding AEF feature set, $T_{desc}$ is a text corpus primarily consisting of comprehensive descriptive semantics (e.g., geographic information such as terrain, landforms, and spatial distributions). This corpus is derived from the FUSAR-GEOVL-1M dataset \cite{fusarklip}, which covers multiple satellite platforms (e.g., GF-3 and QL-1) and multi-resolution imagery, ensuring strong domain generalization capability for the model.

In terms of the training strategy, during this stage we freeze the visual encoder $\theta_v$ and the original weights of the backbone large language model $\theta_{llm}$. The AEF feature-embedding MLP $\theta_{ae}$ and the LoRA adapters $\theta_{\text{lora}}$ injected into the language model are jointly trained during this stage.
The optimization objective for this phase, denoted as $\mathcal{L}_1$, is to maximize the likelihood of the descriptive text $T_{desc}$:
\begin{equation}
\begin{aligned}
&\mathcal{L}_1(\theta_{ae}, \theta_{\text{lora}})
= - \mathbb{E}_{(I, F, T) \in D_1} \Big[
\log P\big(
T \mid E_v(I;\theta_v^{\text{frozen}}),\,\\
&\quad\quad\quad\quad\quad\quad
E_{ae}(F;\theta_{ae});\,
\theta_{llm}^{\text{frozen}}, \,
\theta_{\text{lora}}
\big)
\Big]
\end{aligned}
\label{eq:phase1_loss}
\end{equation}
\begin{equation}
(\theta_{ae}^{*},\, \tilde{\theta}_{\text{lora}})
= \arg\min_{\theta_{ae},\, \theta_{\text{lora}}} \mathcal{L}_1(\theta_{ae}, \theta_{\text{lora}})
\label{eq:phase1_opt}
\end{equation}

This design (as shown in equation \ref{eq:phase1_loss}) forces the MLP layer $\theta_{ae}$, together with the LoRA adapters $\theta_{\text{lora}}$, to learn how to efficiently integrate the multi-source semantic features $(I,F)$ and align the fused representation with the descriptive text $T$. 

\textbf{Stage II: Task Reasoning and LLM Activation}
The core objective of this phase is to activate the model’s analytical capabilities based on the aligned representations so that it can perform specific tasks. We employ the instruction dataset
$D_2 = \{(I_{sar}, F_{ae}, T_{inst}, T_{ans})\}$, where $T_{\text{inst}}$ denotes the task instruction (e.g., localization, classification, counting, and detection), and $T_{\text{ans}}$ is the corresponding ground-truth answer.

In terms of the training strategy, we freeze the visual encoder $\theta_v$, the MLP fusion layer $\theta_{ae}^{*}$ pretrained in Stage One, and the original weights of the large language model backbone $\theta_{llm}$. Starting from the LoRA adapters $\tilde{\theta}_{\text{lora}}$ obtained in Stage I, we continue to update only the LoRA adapter parameters $\theta_{\text{lora}}$ during this stage.

The optimization objective of this stage, denoted as $\mathcal{L}_2$, is to maximize the likelihood of the task answer $T_{\text{ans}}$. The loss function is optimized exclusively over the LoRA parameters $\theta_{\text{lora}}$:
\begin{equation}
\begin{aligned}
&\quad
\mathcal{L}_2(\theta_{\text{lora}}) 
= - \mathbb{E}_{(I, F, T_{\text{inst}}, T_{\text{ans}}) \in D_2} \Big[ 
\log P\big(
T_{\text{ans}} \mid \\
&
E_v(I; \theta_v^{\text{frozen}}),\;
E_{ae}(F; \theta_{ae}^{*}),\;
T_{\text{inst}};\;
\theta_{llm}^{\text{frozen}},\;
\theta_{\text{lora}}
\big)
\Big]
\end{aligned}
\label{eq:phase2_loss_lora}
\end{equation}

\begin{equation}
\theta_{lora}^{*} = \arg\min_{\theta_{lora}} \mathcal{L}_2(\theta_{lora})
\label{eq:phase2_opt_lora}
\end{equation}

Since the input features $E_v(I)$ and $E_{ae}(F;\theta_{ae}^{*})$ have already undergone domain adaptation and cross-modal alignment for SAR data through the optimization of the $\mathcal{L}_1$ stage, the large language model in Stage Two can focus on interpreting the task instruction $T_{\text{inst}}$ and performing complex analysis and inference.

\section{Experiments}
\subsection{Dataset}

The data used in this study are sourced from the FUSAR-GEOVL dataset within FUSAR-KLIP, which retains real latitude and longitude information to satisfy the geolocation requirements of AEF. During initial construction, we selected 10k images and textual descriptions, extracted corresponding AlphaEarth feature vectors based on geographic coordinates, and aligned them to form 10k AEF–image–text triplets. Subsequently, a subset of 2k images containing precise ground-truth target annotations was further selected for downstream task training and evaluation to validate the model's task transfer and generalization performance.

\subsection{Implementation Details}
\begin{figure*}[h]
\centerline{\includegraphics[width=\textwidth]{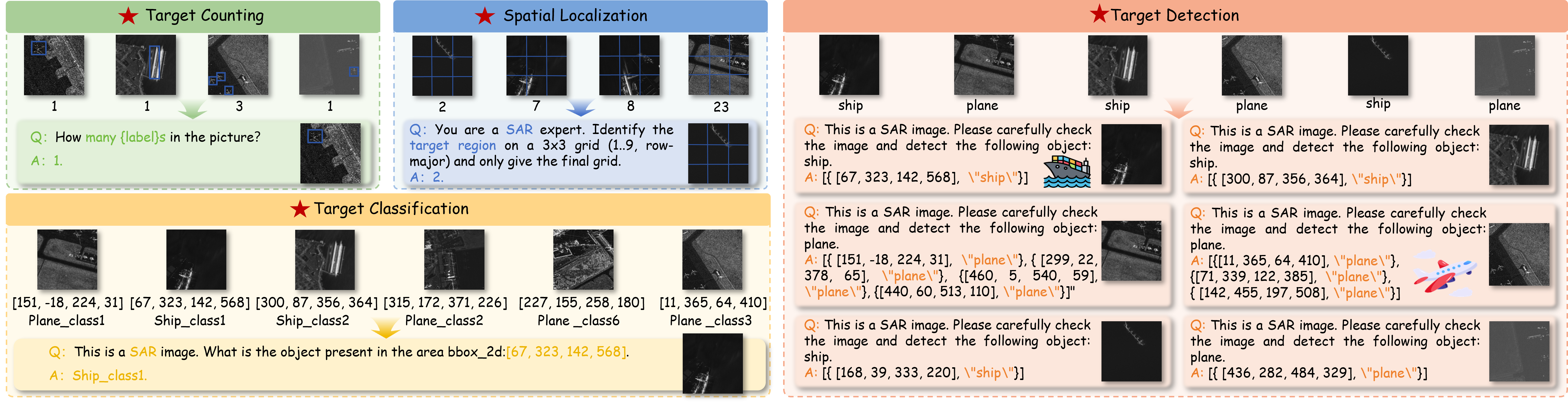}}
\caption{Overview of the four main downstream tasks: Target Counting, Spatial Localization, Target Classification, and Target Detection.}
\label{fig:task_full}
\end{figure*}
All baseline models were fine-tuned in the MS-SWIFT framework under the same parameter settings as FUSAR-GPT for a fair comparison. FUSAR-GPT adopts Qwen2.5-VL-7B as the backbone. To ensure training consistency and stability, the first stage was trained for 30 epochs using a learning rate of $10^{-4}$ to facilitate cross-modal semantic alignment, followed by a second stage of 5 epochs with a learning rate 
of $10^{-5}$ to enable stable task-level
adaptation without disrupting the previously aligned representations. Otherwise, Stage1 and Stage2 share the same system-level training configuration, including Flash-Attention-2, DeepSpeed ZeRO Stage-2, BFloat16 precision, and rank-8 LoRA adapters with LoRA scaling factor $\alpha=32$ applied to all linear layers. Both stages are optimized using AdamW with a warm-up ratio of 0.05, and vLLM is employed to accelerate inference during evaluation.  Since Qwen2.5-VL employs absolute coordinate encoding, we pre-aligned all bounding-box annotations to its unified image resolution. All experiments were conducted on four NVIDIA A100 GPUs.
% 实验设置
% To ensure training consistency and stability, The only difference between the two stages lies in the learning rate setting: \textbf{Stage1} adopts a larger learning rate $1\times10^{-4}$ to facilitate cross-modal semantic alignment, while \textbf{Stage2} uses a smaller learning rate $1\times10^{-5}$ to enable stable task-level adaptation without disrupting the aligned representations.

\subsection{Task}
To comprehensively evaluate the model’s performance in target interpretation, we selected four highly relevant mainstream tasks for testing, as shown in Fig. \ref{fig:task_full}, including target counting, spatial localization, target classification, and target detection. For performance comparison, we employed mainstream vision–language large models as baselines, including models from the Qwen2, Qwen2.5, and Qwen3 series, the LLaVA series, and the InternVL series.

\textbf{Target Counting: }Table \ref{Target counting} reports the performance of various mainstream multimodal models on the SAR counting task. Overall, the accuracy of baseline models mostly falls within the range of 30\% to 40\%, indicating that general vision–language models still encounter notable performance bottlenecks when applied to SAR counting. Moreover, the table shows that increasing model size does not lead to substantial performance gains. For example, within the Qwen3-VL series, the 4-billion-parameter model achieves an accuracy of 45.45\%, while the 8-billion-parameter model drops to 41.41\%, suggesting that scaling up general-purpose models cannot overcome the inherent challenges of SAR imagery, such as strong noise and weak texture. In contrast, FUSAR-GPT attains an accuracy of 52.53\%, exceeding the best baseline by more than 7\%.
\begin{table}[h]
  \centering
  \caption{Performance comparison on the Target Counting and Spatial Localization}
  \label{Target counting}
  \renewcommand{\arraystretch}{1.1}
 \huge
\resizebox{\columnwidth}{!}{
  \begin{tabular}{lcccccc}
    \toprule
    \multirow{2}{*}{\textbf{Model}} &
    \multirow{2}{*}{\textbf{Year}}  &
    \multirow{2}{*}{\textbf{Size}}  &
    \multicolumn{1}{c}{\textbf{Counting}} &
    \multicolumn{3}{c}{\textbf{Spatial Localization}} \\
     &  &  
     & \textbf{@Acc} 
     & \textbf{@Acc100} & \textbf{@Acc50} & \textbf{Top1} \\
    \midrule
    \multirow{2}{*}{Qwen2-VL\cite{qwen2vl}} & \multirow{2}{*}{2024}  & 2B    & 35.86 & 23.23 & 46.97   & 75.25  \\
                    &     & 7B    & 41.41 &  35.35 & 67.68   & 83.33 \\
                  \midrule
    \multirow{2}{*}{Qwen2.5-VL\cite{qwen25vl}} & \multirow{2}{*}{2025} & 3B   & 35.35 & 23.23 &  46.97  & 68.18  \\
                    &      & 7B    &  34.85&  30.81 &  57.07 & 78.28\\
                        \midrule
    \multirow{2}{*}{Qwen3-VL\cite{bai2025qwen3}} & \multirow{2}{*}{2025}  & 4B     &  45.45 &  43.94 &  71.21  & 82.32 \\
    
                    &      & 8B    & 41.41 &  42.93 &  71.72  & 85.86 \\
                     \midrule
    LLaVA-1.5\cite{liu2023improvedllava} &  2023  & 7B    &  40.91  & 35.86  &  63.64  & 84.85\\
   \midrule
    LLaVA-1.6\cite{liu2024llavanext}                &  2024    & 7B    &  44.44 & 39.90 & 70.20   & 85.86  \\
     \midrule
            InternVL-3.5\cite{internvl3_5} & 2025  & 4B  & 40.40 & 39.90 & 70.20 &   83.84 \\
            \midrule
            \rowcolor{blue!5}
           \textbf{FUSAR-GPT} & 2025 &  7B &   \textbf{52.53}& \textbf{52.02} &\textbf{79.29}  &\textbf{91.41}\\
    \bottomrule
  \end{tabular}
  }
\end{table}

\textbf{Spatial Localization: }To evaluate Spatial Localization, we divide each SAR image into a 3$\times$3 grid and instruct the model to identify the target's grid index, assessing basic spatial reasoning and region-alignment. The validation set is diverse: 46.97\% of images contain targets in only one grid, while 22.22\% and 20.71\% contain targets in two and three grids, respectively, and a smaller fraction contain targets across four to eight grids. We employ three metrics: \textbf{Acc@100} (strict exact match), \textbf{Acc@50} (IoU $\ge 0.5$), and \textbf{Top1} (non-empty intersection). As reported in Table \ref {Target counting}, FUSAR-GPT demonstrates substantial advantages across all indicators, achieving 52.02\% on Acc@100, 79.29\% on Acc@50, and 91.41\% on Top1---improving upon the best baseline by 8--12\%. Notably, the significant lead on the Top1 metric indicates FUSAR-GPT's superior stability in multi-target scenarios and a more reliable ability to identify key regions.

% \begin{table}[h]
%   \centering
%   \caption{Spatial localization performance across different VLMs. Results are reported using Acc@100, Acc@50, and Top1. FUSAR-GPT achieves the best accuracy under all metrics.}
%   \label{tab:Spatial localization}
%   \renewcommand{\arraystretch}{1.2}
%   \small
%   \begin{tabular}{llcccc}
%     \toprule
%     \rowcolor{gray!15}
%     \textbf{Model} & \textbf{Size} & \textbf{Acc@100} &  \textbf{Acc@50}  & \textbf{Top1}  \\
%     \midrule
%     \multirow{2}{*}{Qwen2-VL\cite{qwen2vl}} & 2B & 23.23 & 46.97   & 75.25    \\
%                     & 7B  &  35.35 & 67.68   & 83.33  \\
%     \multirow{2}{*}{Qwen2.5-VL\cite{qwen25vl}} & 3B & 23.23 &  46.97  & 68.18    \\
%                     &  7B  &  30.81 &  57.07 & 78.28 \\
%     \multirow{2}{*}{Qwen3-VL\cite{qwen3_blog2025}} & 4B &  43.94 &  71.21  & 82.32   \\
%                     &  8B &  42.93 &  71.72  & 85.86  \\
%     LLaVA-1.5\cite{liu2023improvedllava}                &  7B & 63.64  &  63.64  & 84.85  \\
%     LLaVA-1.6\cite{liu2024llavanext} & 7B & 39.90 & 70.20   & 85.86  \\
                    
%             InternVL-3.5\cite{internvl3_5} & 4B & 39.90 & 70.20 &   83.84 \\
%     \rowcolor{blue!5}
%             \textbf{FUSAR-GPT} & 7B & \textbf{52.02} &\textbf{79.29}  &\textbf{91.41}\\
                    
%     \bottomrule
%   \end{tabular}
% \end{table}
\begin{table*}[t]
\centering
\huge
\caption{Performance comparison on the Target Classification. Results are reported for both coarse-grained categories and fine-grained classes. FUSAR-GPT consistently outperforms Qwen2.5-VL-7B across all category levels.}
\label{tab:Target classification}
\setlength{\tabcolsep}{2pt}
\scriptsize
\resizebox{\textwidth}{!}{
\begin{tabular}{l*{15}{c} ccc}
\toprule
\multirow{2}{*}{\textbf{\textbf{Model}}} 
  & \multicolumn{2}{c}{\textbf{Coarse-grained}} 
  & \multicolumn{13}{c}{\textbf{Fine-grained-plane}}& \multicolumn{3}{c}{\textbf{Fine-grained-ship}} \\
  
\cmidrule(lr){2-3}\cmidrule(lr){4-16}\cmidrule(lr){17-19}
 &\textbf{ Plane }& \textbf{Ship} &\textbf{ P-1} & \textbf{P-2} &\textbf{ P-3} &\textbf{ P-4} &\textbf{ P-5} &\textbf{ P-6} &
  \textbf{ P-7} & \textbf{P-8}&\textbf{ P-9} & \textbf{P-10} & \textbf{P-11} & \textbf{P-12} &\textbf{ P-13} &\textbf{ S-1} & \textbf{S-2} & \textbf{S-3} \\
\midrule
 Qwen2.5-VL-7B\cite{qwen25vl}  & 64.66 & 55.19 & 82.42 & 83.95 & 66.67 & 25.00 & 31.43 & 57.14 &
69.05 & 57.89 & 53.85 & 40.00 & 0.00 & 85.71 & 35.00 & 57.65 & 51.06 & 100.00 \\
\midrule
\rowcolor{blue!5}
\textbf{FUSAR-GPT}  & \textbf{76.85} & \textbf{67.42} & \textbf{85.71} & \textbf{85.19} & \textbf{100} & \textbf{75.00} & \textbf{76.47} & \textbf{66.67} &
\textbf{71.43} & \textbf{36.84} & \textbf{94.44} & \textbf{39.39} & \textbf{71.43} & \textbf{100} & \textbf{90.48} & \textbf{77.65} & \textbf{57.78} & \textbf{66.67} \\
\bottomrule
\end{tabular}
}
\end{table*}

\begin{table*}[!h]
\centering
\renewcommand{\arraystretch}{0.8}
\caption{Performance comparison of Target Detection across different IoU thresholds.}
\label{tab:Target detection}
\setlength{\tabcolsep}{4pt}
\resizebox{\textwidth}{!}{
\begin{tabular}{lccccccccccccccccccc}
\toprule
\multirow{3}{*}{\textbf{Model}} &
\multicolumn{6}{c}{\textbf{IoU = 0.25}} &
\multicolumn{6}{c}{\textbf{IoU = 0.50}} &
\multicolumn{6}{c}{\textbf{IoU = 0.75}} \\
\cmidrule(lr){2-7}\cmidrule(lr){8-13}\cmidrule(lr){14-19}
& \multicolumn{2}{c}{\textbf{All} }& \multicolumn{2}{c}{\textbf{Plane}} & \multicolumn{2}{c}{\textbf{Ship}}
& \multicolumn{2}{c}{\textbf{All} }& \multicolumn{2}{c}{\textbf{Plane}} & \multicolumn{2}{c}{\textbf{Ship}}
& \multicolumn{2}{c}{\textbf{All} }& \multicolumn{2}{c}{\textbf{Plane}} & \multicolumn{2}{c}{\textbf{Ship}} \\
\cmidrule(lr){2-3}\cmidrule(lr){4-5}\cmidrule(lr){6-7}
\cmidrule(lr){8-9}\cmidrule(lr){10-11}\cmidrule(lr){12-13}
\cmidrule(lr){14-15}\cmidrule(lr){16-17}\cmidrule(lr){18-19}
 & \textbf{R} & \textbf{F1}
 & \textbf{R} & \textbf{F1}
 & \textbf{R} & \textbf{F1}
 & \textbf{R} & \textbf{F1}
 & \textbf{R} & \textbf{F1}
 & \textbf{R} & \textbf{F1}
 & \textbf{R} & \textbf{F1}
 & \textbf{R} & \textbf{F1}
 & \textbf{R} & \textbf{F1} \\
\midrule
Qwen2.5-VL-7B\cite{qwen25vl} & 39.9 & 47.1 & 40.3 & 47.5 & 32.3 & 38.5 & 23.5 & 27.7 & 24.3 & 28.7 & 6.5 & 7.7 & 6.2 & 7.3 & 6.3 & 7.5 & 3.2 & 3.8 \\
\midrule
\rowcolor{blue!5}
\textbf{FUSAR-GPT} &
 \textbf{66.4 }& \textbf{74.8} &  \textbf{67.1} & \textbf{75.7} &  \textbf{51.6} & \textbf{57.1} &
 \textbf{52.1} & \textbf{58.7} &  \textbf{53.1} & \textbf{59.8} &  \textbf{32.3} & \textbf{35.7} &
 \textbf{21.4} & \textbf{24.1} &  \textbf{22.0} & \textbf{24.8} &  \textbf{9.7} & \textbf{10.7} \\
\bottomrule
\end{tabular}}
\end{table*}

\textbf{Target Classification:} The model is given explicit bounding boxes and must identify the object category within each box. It is worth noting that the Qwen2-VL, Qwen3-VL, and InternVL series adopt a standardized 0--1000 relative coordinate system, whereas the LLaVA series does not officially support bounding-box coordinates, and prior work has shown that its region-alignment capability is limited\cite{tumu2025referring}\cite{tang2026visual}. Considering both model capability and fairness in bounding-box handling, we therefore conduct comparative experiments only with the Qwen2.5-VL series for this task. From the results in Table \ref{tab:Target classification}, it is evident that FUSAR-GPT achieves substantial gains in both coarse-grained and fine-grained classification tasks. FUSAR-GPT surpasses Qwen2.5-VL-7B by more than 12\% in both coarse-grained categories. In fine-grained classification, the advantage of FUSAR-GPT is even more pronounced.

\textbf{Target Detection: }This task requires the model to detect and output all bounding boxes of a given class from SAR images based on a natural-language category prompt (e.g., “plane”)\cite{li2026co}\cite{ren2025grounding}. As shown in Table \ref{tab:Target detection}, FUSAR-GPT exhibits a substantial advantage under all detection settings. At an IoU threshold of 0.25, the overall F1 score increases by nearly 28\%, rising from 47.1\% to 74.8\%. The F1 score for the plane category improves from 47.5\% to 75.7\%, and for the ship category from 38.5\% to 57.1\%, indicating that the model also demonstrates stronger robustness for small-scale and low-contrast targets. Under a stricter IoU threshold of 0.50, FUSAR-GPT continues to maintain a clear lead.

\begin{figure}[tbp]
\centerline{\includegraphics[width=0.5\textwidth]{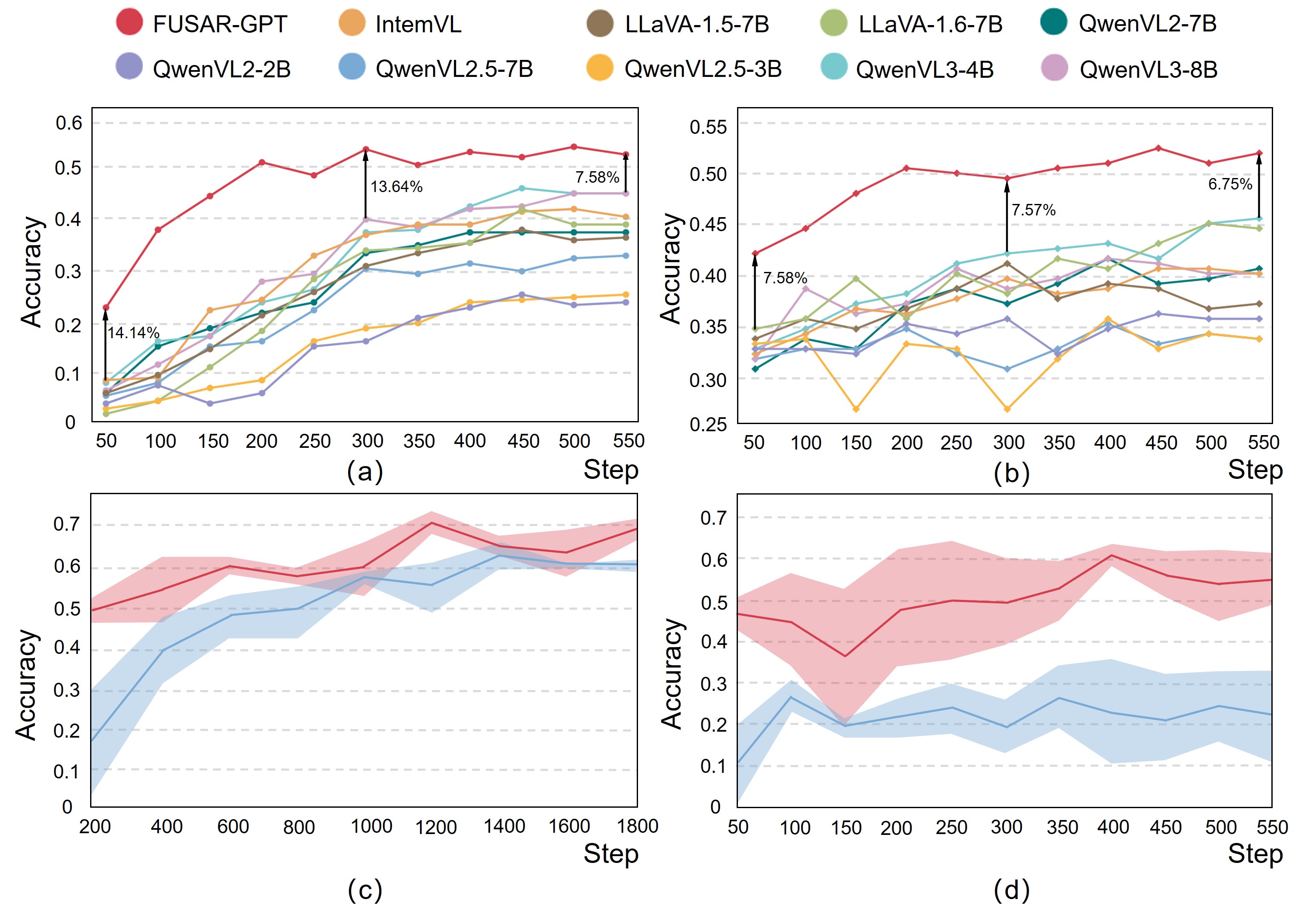}}
\caption{Performance of different training stages across downstream tasks.
a: Target Counting;
b: Spatial Localization;
c: Target Classification;
d: Target Detection.}
\label{fig:all_results}
\end{figure}

As shown in  Fig.\ref{fig:all_results}, we further analyze the accuracy dynamics throughout the training process. It is clearly observable that FUSAR-GPT demonstrates performance far exceeding all baseline models from the initial stages of training, consistently maintaining a significant lead. This indicates that FUSAR-GPT not only achieves superior final performance but also exhibits faster convergence and higher data efficiency, likely attributable to the robust prior knowledge injected during SFT Stage 1.

\section{Conclusion}

To address the challenge that general VLMs exhibit limited performance in the SAR domain due to modality gaps and feature sparsity, we propose FUSAR-GPT. Our approach introduces AEF features as multi-source geospatial priors that leverage the stable nature of large-scale geographic semantics, and dynamically perform semantic compensation for sparse representations in SAR imagery through spatiotemporal anchors. To enable efficient fusion, we design the TLM module, which applies affine transformations to SAR visual tokens in a modulation manner. In addition, the proposed two-stage decoupled fine-tuning framework separates the model’s knowledge injection and task execution processes at the parameter level. Experimental results show that FUSAR-GPT achieves competitive performance across multiple SAR vision–language downstream tasks.

\section{Acknowledgements}
This work was supported  by the National Natural Science
Foundation of China under Grant 62271153.
{
    \small
    \bibliographystyle{ieeenat_fullname}
    \bibliography{main}
}

\clearpage
\setcounter{page}{1}
\maketitlesupplementary

\section{The specificity of SAR images}
\begin{figure}[htbp]
\centerline{\includegraphics[width=0.5\textwidth]{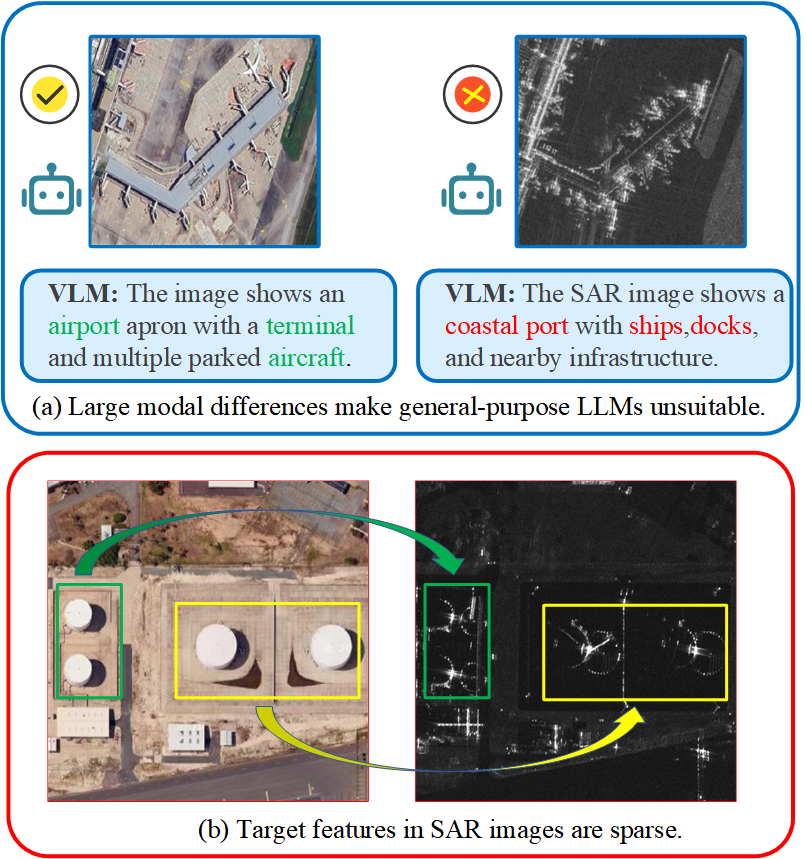}}
\caption{Challenges in developing SAR visual language models.}
\label{fig2_mul}
\end{figure}

As illustrated in Fig. \ref{fig2_mul}, SAR imagery exhibits inherent limitations that constrain visual–semantic understanding. First, the large modality gap between optical and SAR images leads to systematic misinterpretation: optical models rely on color, texture, and shading, whereas SAR captures microwave backscatter dominated by geometric and dielectric responses. Consequently, general-purpose VLMs often map SAR scattering patterns to incorrect optical concepts, such as hallucinating roads or vehicles. Second, SAR imagery is intrinsically sparse—strong scatterers form a small number of saturated bright points, while extensive dark regions contain weak but semantically relevant signals that lack explicit structural cues. This highly polarized intensity distribution causes model attention to collapse onto a few bright pixels, suppressing contextual information embedded in the dark background. Together, these modality discrepancies and sparsity characteristics highlight the fundamental limitations of applying standard VLMs to SAR imagery.

\section{Additional Experiments}
% Thank you for the reviewers’ suggestion. In response to Reviewer, we include additional results for SAR-CLIP and SARLANG-1M. For Reviewer and Reviewer Q4, we further evaluate multiple remote sensing vision–language models on the several SAR understanding tasks. Specifically, we report corpus-level captioning performance in Table\ref{tab:fusar_caption_results}.
% In addition, we conduct quantitative comparisons on three representative SAR tasks in Table\ref{tab:sar_multitask_results}. 

Table \ref{tab:sar_multitask_results} presents an extended comparison on three SAR understanding tasks with additional baselines. FUSAR-GPT consistently outperforms all competing methods by a significant margin across counting, grid-based localization, and classification tasks. In particular, it achieves substantial improvements in counting accuracy and grid prediction, highlighting its strong spatial reasoning capability. Compared with recent remote sensing–oriented models such as SAR-CLIP, SAR-JEPA, and SkyCLIP, our method demonstrates superior cross-task generalization, indicating that the combination of SFT1 pre-alignment and TLM-based feature fusion enables a more unified and robust representation for diverse SAR understanding tasks.
\begin{table}[h]
\centering
\caption{Performance comparison on SAR understanding tasks. B and L denote ViT-Base and ViT-Large backbones, respectively.}

\label{tab:sar_multitask_results}
\footnotesize
\setlength{\tabcolsep}{2pt}   % 关键：压缩列间距
\renewcommand{\arraystretch}{1.0}
\begin{tabular}{l c c c c c}
\hline
\multirow{2}{*}{\textbf{Model}}
& \textbf{Count}
& \multicolumn{3}{c}{\textbf{Grid}} 
& \textbf{Classification} \\
\cline{3-5}
& \textbf{@Acc}
& \textbf{Acc@100}
& \textbf{Acc@50}
& \textbf{Top1}
& \textbf{@Acc} \\
\hline
BAN-L\cite{li2024new}            & 21.21 & 12.63 & 23.74    & 37.37    & 57.83 \\
ChangeCLIP-B\cite{dong2024changeclip}     & 27.78 & 14.14 & 30.81    & 43.43    & 72.59 \\
GeoRSCLIP-B\cite{zhang2024rs5m}      & 30.30 & 10.61 & 20.71    & 37.88    & 68.98 \\
Prithvi-B\cite{blumenstiel2024multi}        & 21.21 &  8.08 & 17.17    & 28.79    & 40.66 \\
RemoteCLIP-L\cite{liu2024remoteclip}     & 31.31 & 11.62 & 21.71    & 37.88    & 67.92 \\
SAR-CLIP-B\cite{JIANG202617}       & 31.82 & 29.80 & 44.95    & 65.66    & 69.43 \\
SAR-JEPA-B\cite{li2024predicting}      & 28.28 & 31.82 & 57.58    & 68.69    & 58.89 \\
SkyCLIP-L\cite{Wang_Prabha_Huang_Wu_Rajagopal_2024}     & 21.21 & 27.78 & 49.50    & 60.61    & 64.16 \\
SARLANG-1M\cite{wei2026sarlang}  & 46.97 & 36.36 & 67.17    &   83.33  & 64.61 \\
BITA-L\cite{yang2024bootstrapping}            & 21.21 & 10.61 & 18.18    &   35.35  & 60.99 \\
GeoChat-L\cite{kuckreja2024geochat}           & 31.82 & 9.60 & 16.67    &   39.90  &  69.73 \\
VHM-L\cite{pang2025vhm}           & 17.17 & 7.58 & 14.65    &   30.81  & 68.83 \\
\rowcolor{blue!10}
\textbf{FUSAR-GPT}
              & \textbf{52.53}
              & \textbf{52.02}
              & \textbf{79.29}
              & \textbf{91.41}
              & \textbf{74.04} \\
\hline
\end{tabular}
% \vspace{-15pt}
\end{table}

\begin{figure}[!h]
\centerline{\includegraphics[width=0.5\textwidth]{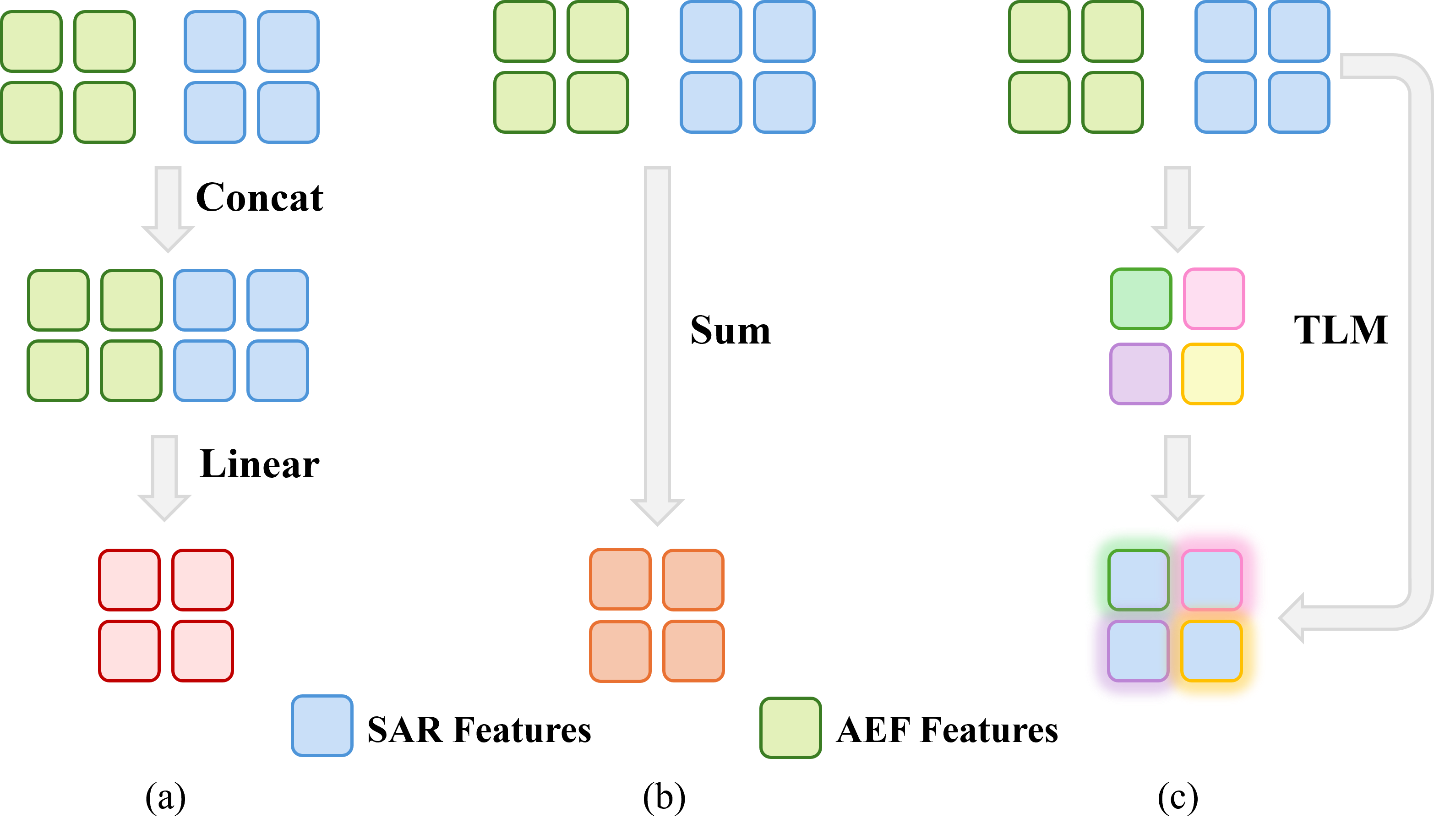}}
\caption{Different Fusion Strategies}
\label{sul_diff}
\end{figure}

We conduct comparisons on the Target Counting task with proprietary large models, including ChatGPT-5.2 and Gemini-3. As shown in Fig.\ref{sul_diff}, we explore different strategies for integrating AEF features. Table \ref{tab:counting_other_model1} shows that our method consistently outperforms these strong baselines, while the proposed TLM-based fusion achieves the best performance among all integration schemes. These findings further validate the effectiveness of the TLM module in leveraging AEF priors for improved SAR understanding.

\begin{table}[h]
\setlength{\tabcolsep}{2pt}

\renewcommand{\arraystretch}{0.5}
\caption{Models evaluated on Target Counting task.}
\label{tab:counting_other_model1}
\centering
\footnotesize
\resizebox{\columnwidth}{!}{
\begin{tabular}{c c c c c c c}
\toprule
Model & Base & TLM & Sum & Concat & ChatGPT-5.2 & Gemini-3 \\
\midrule
Accuracy & 34.85 & 52.53  & 36.36 & 37.88 & 18.18 & 30.3\\
\bottomrule
\end{tabular}
}
\end{table}

\begin{table*}[htbp]
\centering
\caption{Corpus-level captioning performance.}
% 在导言区设置（影响所有表格）
\label{tab:fusar_caption_results}
% \scriptsize
% \setlength{\tabcolsep}{3pt}
% \renewcommand{\arraystretch}{1.0}
\begin{tabular}{lccccccc}
\hline
\textbf{Model} & \textbf{Bleu-1} & \textbf{Bleu-2} & \textbf{Bleu-3} & \textbf{Bleu-4} & \textbf{CIDEr} & \textbf{METEOR} & \textbf{SPICE} \\
\hline
LLaVA-1.5-7B\cite{liu2023improvedllava}        & 73.54 & 67.20 & 61.34 & 55.59 & 81.97 & 42.13 & 84.14 \\
LLaVA-1.6-7B\cite{liu2024llavanext}        & 75.03 & 69.08 & 63.35 & 57.83 & 93.63 & 43.13 & 84.94 \\
InternVL-3.5-4B\cite{internvl3_5}     & 76.76 & 70.76 & 64.99 & 59.36 & 116.21 & 42.34 & 84.36 \\
Qwen2-VL-3B\cite{qwen2vl}      & 72.11 & 65.80 & 59.80 & 54.11 & 91.34 & 41.20 & 83.07 \\
Qwen2-VL-7B\cite{qwen2vl}      & 76.11 & 70.20 & 64.57 & 59.13 & 109.11 & 42.52 & 84.62 \\
Qwen2.5-VL-3B\cite{qwen25vl}    & 61.56 & 56.27 & 51.17 & 46.24 & 81.68 & 35.88 & 81.22 \\
Qwen2.5-VL-7B\cite{qwen25vl}    & 66.71 & 61.39 & 56.22 & 51.19 & 91.68 & 37.95 & 83.02 \\
Qwen3-VL-4B\cite{bai2025qwen3}      & 77.64 & 71.61 & 65.87 & 60.31 & 116.20 & 43.13 & 85.45 \\
Qwen3-VL-8B\cite{bai2025qwen3}      & 77.41 & 71.49 & 65.84 & 60.36 & 110.79 & 43.09 & 85.71 \\
SARLANG-1M\cite{wei2026sarlang}       & 74.56 &   69.84  & 65.17 &  60.61  & 149.60  & 43.16 & 88.12  \\
BAN(ViT-L)\cite{li2024new}                & 58.08 & 51.39 & 44.99 & 38.79 & 14.08 & 32.79 & 56.99 \\
ChangeCLIP(ViT-B)\cite{dong2024changeclip}         & 59.95 & 53.54 & 47.58 & 41.52 & 16.40 & 34.12 & 57.29 \\
GeoRSCLIP(ViT-B)\cite{zhang2024rs5m}          & 59.01 & 52.62 & 46.51 & 40.35 & 16.63 & 33.20 & 57.49 \\
Prithvi(ViT-B)\cite{blumenstiel2024multi}            & 60.05 & 53.33 & 46.83 & 40.37 & 11.72 & 33.37 & 61.81 \\
RemoteCLIP(ViT-L)\cite{liu2024remoteclip}         & 61.12 & 54.54 & 48.10 & 41.64 & 17.09 & 34.66 & 59.19 \\
SAR-CLIP(ViT-B)\cite{JIANG202617}           & 59.91 & 53.41 & 47.05 & 40.73 & 13.78 & 34.06 & 61.13 \\
SAR-JEPA(ViT-B)\cite{li2024predicting}           & 57.11 & 52.65 & 44.79 & 37.21 & 10.03 & 32.11 & 58.44 \\
SkyCLIP(ViT-L)\cite{Wang_Prabha_Huang_Wu_Rajagopal_2024}          & 60.77 & 54.19 & 47.95 & 41.69 & 14.91 & 34.74 & 57.81 \\
BITA (ViT-L)\cite{yang2024bootstrapping}              & 55.37 & 44.72 & 37.32 & 30.72 & 2.26 & 28.33 & 56.13 \\
GeoChat(ViT-L)\cite{kuckreja2024geochat}            & 54.04 & 43.55 & 36.10 & 29.67 & 2.19 & 26.11 & 57.66 \\
VHM(ViT-L)\cite{pang2025vhm}                & 50.29 & 41.92 & 35.36 & 29.25 & 3.14 & 27.27 & 58.91 \\
\rowcolor{blue!10}
\textbf{FUSAR-GPT} 
                  & \textbf{77.92} & \textbf{73.21} & \textbf{68.62} & \textbf{64.01} 
                  & \textbf{160.21} & \textbf{45.60} & \textbf{89.29} \\
\hline
\end{tabular}
% \vspace{-10pt}
\end{table*}

At IoU = 0.5, we conduct a preliminary comparison with a representative remote sensing detector, R3Det, on a SAR object detection task. Although our approach is not specifically designed for detection, it demonstrates competitive performance. These results provide an initial indication of the potential of our framework for SAR detection tasks. We leave a more comprehensive evaluation and further exploration of detection-oriented adaptations to future work.
\begin{table}[h]

\renewcommand{\arraystretch}{0.5}
\caption{Models evaluated on Target Detection task.}
\label{tab:target_det_r3_1}
\centering
\footnotesize
% \resizebox{\columnwidth}{!}{
\begin{tabular}{c c c c }
\toprule
Model & P & R & F1 \\
\midrule

R3Det\cite{yang2021r3det} & 62.77 & 44.73   &  52.20 \\
\rowcolor{blue!5}
FUSAR-GPT & 66.79 & 52.10 & 58.70 \\
\bottomrule
\end{tabular}
% }
\end{table}

% To validate the generalization capability of our model, we extend the evaluation to a corpus-level captioning task, as shown in Table \ref{tab:fusar_caption_results}. Compared with both general-domain VLMs  and remote sensing–specific models, FUSAR-GPT achieves the best performance across all metrics, including BLEU, CIDEr, METEOR, and SPICE. Notably, it surpasses the strongest baseline by a clear margin in CIDEr and BLEU-4, indicating its superior ability to generate semantically rich and structurally accurate descriptions for SAR imagery. These results demonstrate that the proposed framework effectively captures both low-level scattering characteristics and high-level semantic information, leading to more faithful and informative caption generation.
To validate the generalization capability of our model, we further extend the evaluation to a corpus-level captioning task, as shown in Table \ref{tab:fusar_caption_results}. Compared with both general-domain VLMs and remote sensing–specific models, FUSAR-GPT achieves the best performance across all metrics, including BLEU, CIDEr, METEOR, and SPICE. Notably, it surpasses the strongest baseline by a clear margin in CIDEr and BLEU-4, indicating its superior ability to generate semantically rich and structurally accurate descriptions for SAR imagery.

% Beyond overall metric improvements, the consistent gains across diverse evaluation criteria suggest that our model effectively balances lexical diversity and semantic fidelity, while maintaining structural coherence in long-form descriptions. This can be attributed to the integration of AEF-based geospatial priors and the two-stage training paradigm, which jointly enhance both representation robustness and language alignment.

% These results demonstrate that the proposed framework captures both low-level scattering characteristics and high-level semantic information, leading to more faithful, informative, and context-aware caption generation.
\section{Ablation Experiment}
\begin{table}[h]
  \centering
  \caption{Ablation results on the target counting task.
Each component:SFT1, SFT2, and TLM independently contributes to performance improvement, while combining all modules yields the highest accuracy.}
\small
  \label{tab:Ablation Experiment}
  \begin{tabular}{lcccc}
    \toprule
 \rowcolor{gray!15}
    \textbf{Model} & \textbf{SFT1} & \textbf{SFT2} & \textbf{TLM} & \textbf{ACC}  \\
    \midrule
    BaseModel & \xmark & \xmark & \xmark & -  \\
    BaseModel\raisebox{-0.1ex}{\scriptsize (+SFT2)} & \xmark & \cmark &\xmark & 34.85  \\
    BaseModel\raisebox{-0.1ex}{\scriptsize (+SFT1+SFT2)} & \cmark & \cmark &\xmark & 36.36 \\
   BaseModel\raisebox{-0.1ex}{\scriptsize (+SFT2+TLM)} & \xmark& \cmark &\cmark & 41.92  \\
    \rowcolor{blue!5}
    \textbf{FUSAR-GPT} & \cmark & \cmark &\cmark & \textbf{52.53}  \\
    \bottomrule
  \end{tabular}%
\end{table}

 \begin{figure*}[!h]
\centerline{\includegraphics[width=\textwidth]{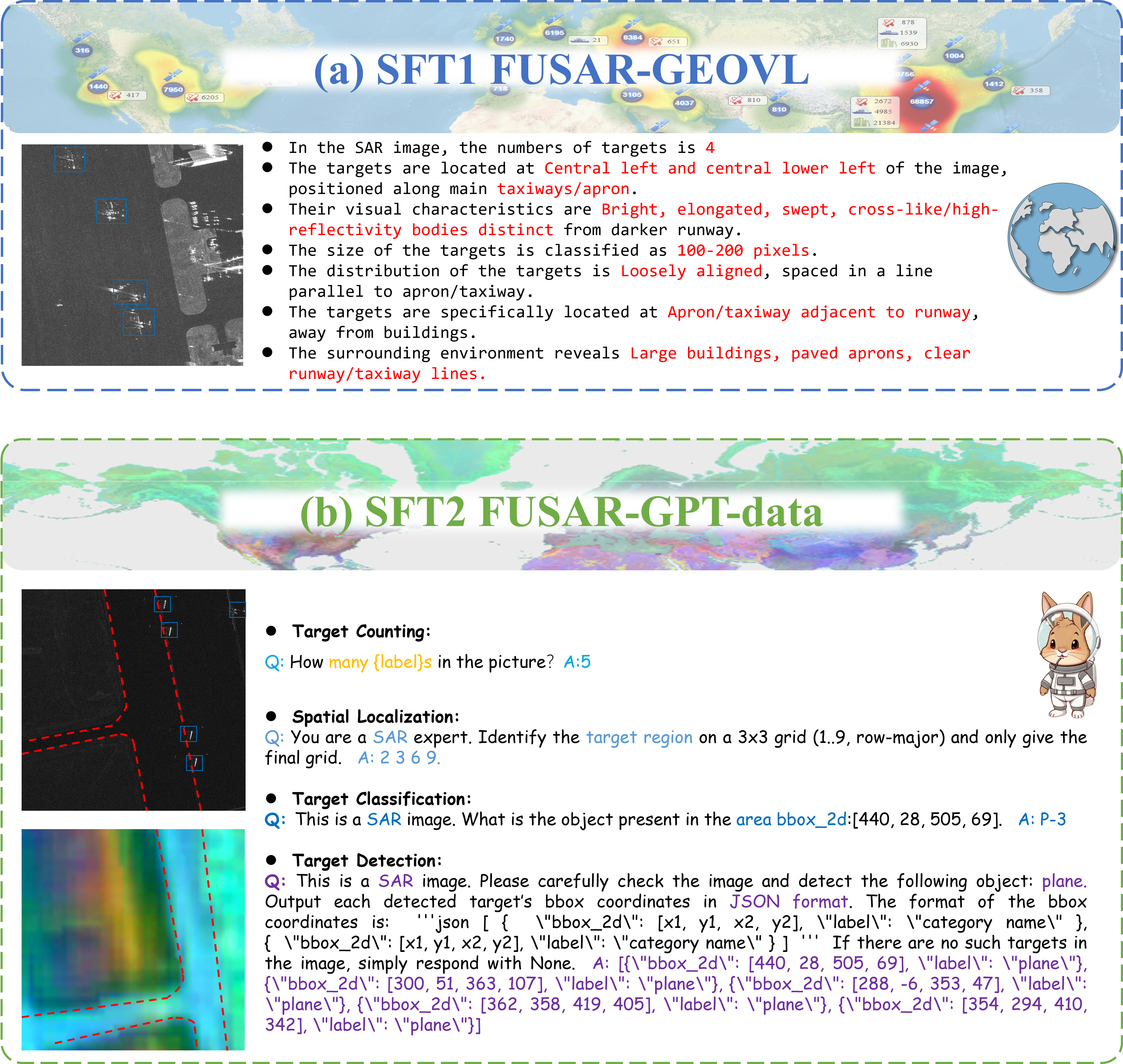}}
\caption{(a) shows the data characteristics of FUSAR-GEOVL, (b) shows the data characteristics of FUSAR-GPT.}
\label{2008}
\end{figure*}
To supplement the main text, we conduct detailed ablation studies on the target counting task to evaluate the contribution of the first-stage supervised fine-tuning (SFT1), which aims to inject SAR-specific knowledge into the model. Unlike SFT2, which directly targets downstream tasks, SFT1 enhances the model’s basic understanding of SAR imaging characteristics and domain patterns, providing a more informative initialization.

As shown in Table \ref{tab:Ablation Experiment}, the Qwen2.5-VL baseline using only SFT2 achieves a performance of 34.85\%, revealing a significant bottleneck in SAR understanding. Incorporating the TLM module to fuse AEF features improves the performance to 41.92\%, demonstrating the effectiveness of AEF geographical priors as an independent knowledge source. Further introducing SFT1 yields an additional gain of approximately 10 percentage points, ultimately boosting the full model to 52.53\%.
These results highlight a strong synergy between SFT1 pre-alignment and TLM feature fusion: SFT1 establishes a superior semantic alignment foundation, enabling TLM to more effectively exploit AEF priors, thereby strengthening domain awareness and achieving optimal downstream performance.
% We conduct detailed ablation studies on the target counting task. Table \ref{tab:Ablation Experiment} presents the ablation study results for key components of FUSAR-GPT. The QWen2.5VL baseline model using only SFT2 achieves a performance of only 34.85\%, revealing its significant bottleneck in SAR tasks. Introducing the TLM module to fuse AEF features significantly improves the performance to 41.92\%, highlighting the significant value of AEF geographical priors as an independent knowledge source. Building on this, the SFT1 pre-alignment stage provides a nearly equivalent additional gain, ultimately boosting the performance of the complete model to 52.23\%. This result clearly demonstrates a strong synergy between SFT1 pre-alignment and TLM feature fusion: SFT1 provides a superior semantic alignment foundation, enabling TLM to more fully leverage the potential of AEF priors, thus jointly achieving optimal performance.

\section{Data Description}
 % To illustrate the differences between the two training stages, we present examples of the data used in SFT1 and SFT2, as shown in Fig. \ref{2008}. The SFT1 stage leverages data from FUSAR-GEO-VL, which provides SAR–geospatial paired descriptions containing detailed structural cues, spatial relations, and domain-specific semantic attributes. This stage injects fundamental SAR knowledge into the model. In contrast, SFT2 uses downstream FUSAR-GPT-data, which focuses on instruction-style supervision for tasks such as target counting and spatial localization. Together, these two complementary datasets enable the model to first acquire broad SAR domain understanding and then adapt to fine-grained task objectives.

As shown in Fig.\ref{2008}, the SFT1 stage is built upon the FUSAR-GEOVL dataset, which provides richly structured and multi-dimensional semantic information that substantially enhances the model’s understanding of SAR imagery. Unlike conventional SAR datasets that offer only category labels or sparse annotations, FUSAR-GEOVL incorporates geographic metadata, multi-scale spatial context, landform descriptions, regional functional cues, and detailed target-level attributes—covering scattering characteristics, structural patterns, spatial layout relations, and environmental semantics. This information-rich annotation design, grounded in SAR physical imaging principles and global-to-local cognitive reasoning, supplies the model with a broad and coherent knowledge base. Through SFT1, the model internalizes these diverse information dimensions, enabling stronger perception of SAR-specific structures, better discrimination of subtle scattering variations, and improved reasoning over spatial and contextual relationships. As a result, SFT1 serves as an effective knowledge injection phase that significantly strengthens the model’s domain awareness before downstream task tuning.

\end{document}